%% file: main_arxiv.tex
\documentclass[10pt,twocolumn,letterpaper]{article}
\usepackage{cvpr}      
\usepackage{multirow, float, booktabs, boldline, hhline}
\usepackage{amsmath}

\input{math}
\input{preamble}
\definecolor{cvprblue}{rgb}{0.21,0.49,0.74}
\usepackage[pagebackref,breaklinks,colorlinks,allcolors=cvprblue]{hyperref}
\usepackage{flushend}
\usepackage{balance}
\def\paperID{978} 
\def\confName{CVPR}
\def\confYear{2026}

\title{Fine-Grained Zero-Shot Learning with Attribute-Centric Representations}
\author{Zhi Chen$^{1}$ \quad Jingcai Guo$^{2}$  \quad Taotao Cai $^1$ \quad Yuxiang Cai$^3$ \\ $^1$University of Southern Queensland   $^2$The Hong Kong Polytechnic University   $^3$ Zhejiang University\\
uqzhichen@gmail.com}

\begin{document}
\maketitle
\begin{abstract}
Recognizing unseen fine-grained categories demands a model that can distinguish subtle visual differences. This is typically achieved by transferring visual-attribute relationships from seen classes to unseen classes. The core challenge is attribute entanglement, where conventional models collapse distinct attributes like color, shape, and texture into a single visual embedding. This causes interference that masks these critical distinctions. The post-hoc solutions of previous work are insufficient, as they operate on representations that are already mixed. We propose a zero-shot learning framework that learns Attribute-Centric Representations (ACR) to tackle this problem by imposing attribute disentanglement during representation learning. ACR is achieved with two mixture-of-experts components, including Mixture of Patch Experts (MoPE) and Mixture of Attribute Experts (MoAE). First, MoPE is inserted into the transformer using a dual-level routing mechanism to conditionally dispatch image patches to specialized experts. This ensures coherent attribute families are processed by dedicated experts. Finally, the MoAE head projects these expert-refined features into sparse, part-aware attribute maps for robust zero-shot classification. On zero-shot learning benchmark datasets CUB, AwA2, and SUN, our ACR achieves consistent state-of-the-art results.

\end{abstract}

\section{Introduction}
\label{sec:intro}
\begin{figure}[t]
  \centering
  \includegraphics[width=0.47\textwidth]{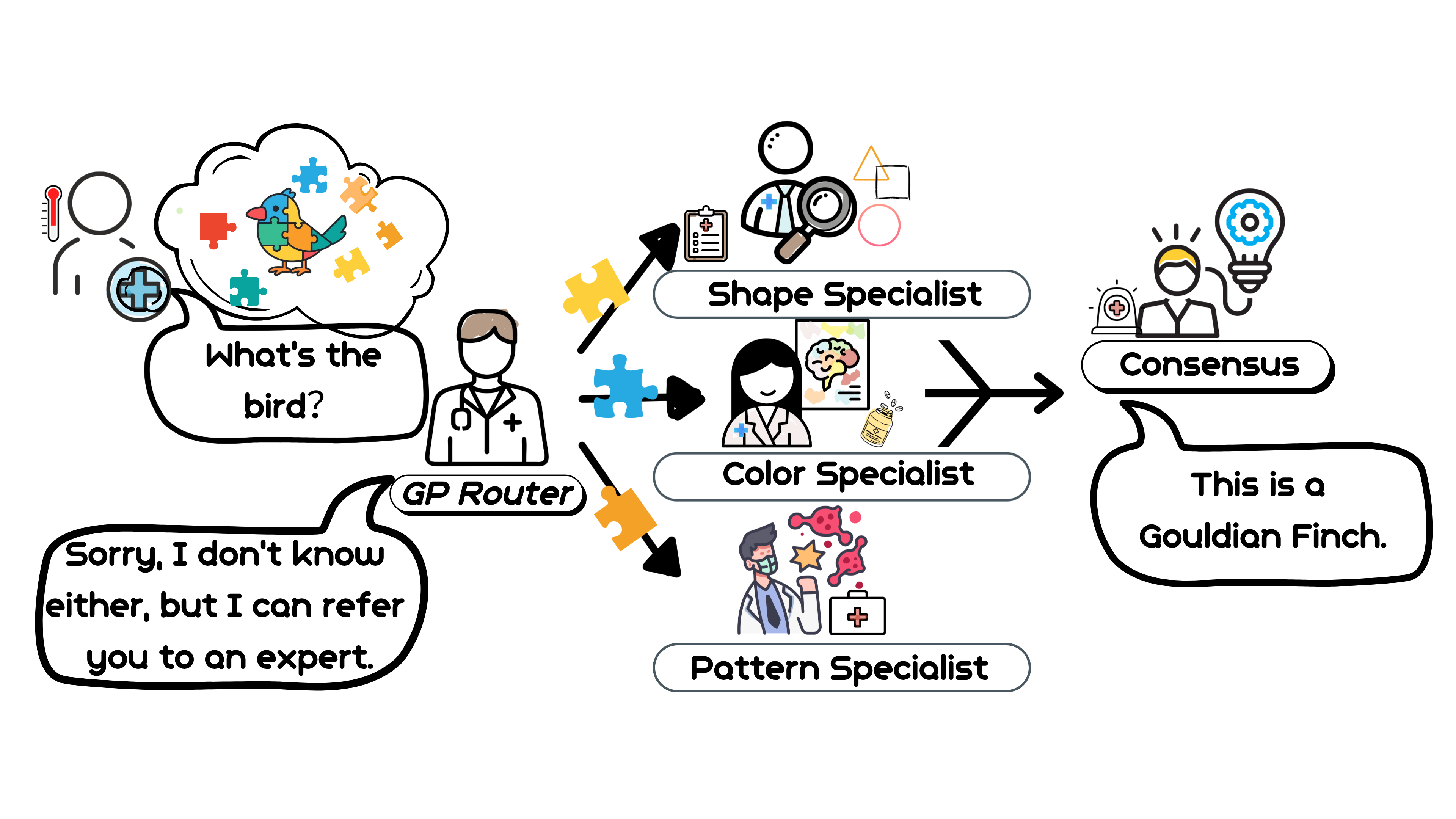}
  \vspace{-15pt}
  \caption{Overall idea of the attribute-centric representations for fine-grained zero-shot learning.
Like a general practitioner (GP) who triages a patient to the right specialists, our GP router examines coarse visual appearance and routes each image patch to the most relevant attribute experts. Each specialist analyzes a coherent attribute group, and their consensus forms the final recognition output, which enables interpretable, attribute-centric reasoning.}
  \label{fig:intro}
\end{figure}

Zero-shot learning (ZSL) ~\cite{palatucci2009zero,xian2017zero,kodirov2017semantic,tong2019hierarchical,han2020learning,chen2021free,chen2021semantics,li2021generalized,chen2024progressive,chen2020canzsl,chen2020rethinking,chen2023zero} aims to recognize unseen categories by transferring knowledge from seen classes. This transfer is typically mediated by human-defined semantic attributes or textual descriptions. The challenge is amplified in fine-grained regimes, where categories differ by only subtle visual parts. A model must distinguish between highly similar instances, such as a Gouldian Finch and a Pine Grosbeak, based on tiny variations in features like bill shape or breast pattern.

Conventional approaches, including large vision language models like CLIP~\cite{radford2021learning}, treat the visual branch as a monolithic feature extractor. This prevailing design forces the model to collapse heterogeneous visual features, such as color, shape, and texture, into a dense embedding. In the fine-grained domain, this approach is detrimental. It inevitably creates attribute entanglement and semantic interference. These challenges cause the model to amplify dominant patterns, while at the same time suppressing the subtle but important patterns for accurate recognition.

The majority of prior work attempts to address this fundamental problem after feature extraction. These approaches implement post-hoc fixes, such as \textit{recalibrating classifier heads}~\cite{lampert2013attribute,liu2018generalized}, \textit{synthesizing unseen features}~\cite{xian2018feature,zhu2018generative,chen2021semantics}, or \textit{projecting into attribute subspaces}~\cite{xu2020attribute,chen2024progressive}. Despite their effectiveness, they must operate on an encoder that has already mixed attribute representations across its layers. Even within Vision Transformers (ViTs) ~\cite{dosovitskiy2020image}, where patch structure is exposed, attribute semantics diffuse~\cite{qin2022devil} as tokens repeatedly exchange information without an explicit inductive bias for attribute-wise organization. Any late-stage pruning or reweighting is insufficient, as it operates on features in which important attribute evidence has already been fused and corrupted.

Therefore, our motivation is that in ZSL, \textbf{disentanglement must be imposed inside representation modelling}. Attributes should not merely supervise the classifier. Instead, they should actively determine how information flows through the network. As illustrated in Figure \ref{fig:intro}, an effective model should function less like a single, overwhelmed extractor and more like a General Practitioner (GP) Router. This GP would first examine coarse visual features, then route each image patch to the most relevant attribute specialists: a ``Shape Specialist," a ``Color Specialist," or a ``Pattern Specialist." The consensus of these specialists would then present the final, interpretable identification.

To achieve this paradigm, we propose a transformer framework that learns Attribute-Centric Representations (ACR) by utilizing a semantically grounded mixture of experts architecture. ACR imposes disentanglement during representation learning via two core components, including a Mixture of Patch Experts (MoPE) and a Mixture of Attribute Experts (MoAE) head. 
Specifically, we insert lightweight MoPE adapters, implemented as LoRA modules for efficiency, between the self-attention and feed-forward layers of a ViT. These experts are controlled by a novel dual-level routing mechanism. First, an instance-level router activates a small set of attribute groups relevant to the whole image, and second, a token-level router assigns individual patches to their most pertinent active experts.
This conditional routing allocates capacity specifically to coherent attribute families before interference can arise, allowing the model to scale its knowledge capacity without increasing per-sample computational cost. At the output, the MoAE head projects the routed patch features into sparse, part-aware attribute maps for zero-shot classification. However, stabilizing conditional routing is non-trivial. To prevent expert collapse, where a few experts dominate traffic, we introduce a load-balancing objective. To ensure specialists remain coherent, we use a cross-layer consistency term that penalizes erratic routing paths. Finally, an entropy-based diversity loss promotes expert exploration early in training. Unlike generic MoE designs, our experts are explicitly attribute-centric and supervised semantically at the output, which grounds their specialization directly in the ZSL objective.
Our contributions are threefold:

\begin{itemize}
\item 
We propose ACR, a novel ZSL framework that enforces attribute wise disentanglement during representation learning using a semantically grounded mixture of experts architecture, moving beyond post hoc corrections.
\item 
We introduce a novel dual-level (instance and patch) routing mechanism and an effective training strategy, incorporating load balancing, cross-layer consistency, and diversity losses, to achieve stable and interpretable expert specialization.
\item 
ACR establishes new state of the art performance on the CUB, AwA2, and SUN benchmarks in both ZSL and Generalized ZSL settings, while simultaneously producing part aware, interpretable attribute maps.
\end{itemize}

\section{Related Work}
\label{sec:related_work}
\noindent \textbf{Zero-Shot Learning (ZSL).} 
Recent progress in zero-shot learning \cite{verma2020meta,han2021contrastive,yu2020episode,liu2020attribute,xie2020region,cheng2023hybrid,liu2023progressive,vyas2020leveraging,su2022distinguishing,gowda2023synthetic,kim2022semantic,zhao2023gimlet,feng2020transfer,chen2023evolving,chen2024causal,liu2021goal,chen2021mitigating,chen2021entropy,chengsmflow,guo2024fine} has unfolded along three complementary lines. Generative feature synthesis methods first showed that a model can hallucinate discriminative visual representations for unseen categories and then train a conventional classifier on them. The f-CLSWGAN of \cite{xian2018feature} pioneered this idea and remains a strong baseline for generalized ZSL. Transformer-based attribute localisation has since emerged: TransZero++ \cite{chen2022transzero++} couples dual attribute-guided transformers to align visual patches with semantic descriptions and now sets the state of the art on CUB, AWA2 and SUN, while the very recent Progressive Semantic-Guided ViT refines these cues layer-by-layer to further boost transfer accuracy \cite{chen2024progressive}. The third strand leverages large vision–language pre-training. CLIP \cite{radford2021learning} demonstrated that contrastive alignment of 400 M image–text pairs yields strong zero-shot recognition “for free”, prompting a wave of prompt-learning adaptations that fine-tune only textual templates. CoOp \cite{zhou2022learning} introduced learnable context vectors to tailor CLIP to downstream datasets, and CoCoOp \cite{zhou2022conditional} added instance-conditioned prompts to avoid over-fitting to seen classes and generalise better to unseen ones. These advances in feature generation, attribute-aware transformers, and prompt-driven vision–language models, collectively define the frontier that our expert-routed, part-disentangled approach seeks to unify and extend.

\noindent \textbf{Mixture of Experts.} Mixture of Experts (MoE) \cite{jacobs1991adaptive,shazeer2017outrageously} architectures decompose complex tasks into simpler ones using a learned router to activate a handful of specialist sub-networks called experts. Early work \cite{shazeer2017outrageously} proposed the sparsely-gated MoE layer and showed that conditional computation could scale language models by three orders of magnitude without slowing training. 
Subsequent engineering refinements—most notably the Switch Transformer \cite{fedus2022switch}, which simplifies routing to a single active expert per token and adds stabilising regularisers—pushed MoE to the trillion-parameter regime and demonstrated strong gains in multilingual NLP. 
In computer vision, \cite{riquelme2021scaling} transplanted the idea into the Vision Transformer, yielding V-MoE, a sparse ViT that matches the accuracy of very large dense backbones while using roughly half the test-time compute and allowing adaptive per-image routing. Systems work such as Tutel \cite{hwang2023tutel} has since addressed the communication and load-balancing bottlenecks that arise when MoE layers are distributed over hundreds of GPUs, making sparse vision models practical at scale. 
Our approach builds on these insights but shifts the focus from class-level prediction to attribute-level reasoning, where each expert is encouraged to specialize in a group of attributes such as head colour and tail pattern, and consistency constraints align the routing decisions made at different transformer depths.

\begin{figure*}[t]
  \centering
  \includegraphics[width=1.0\textwidth]{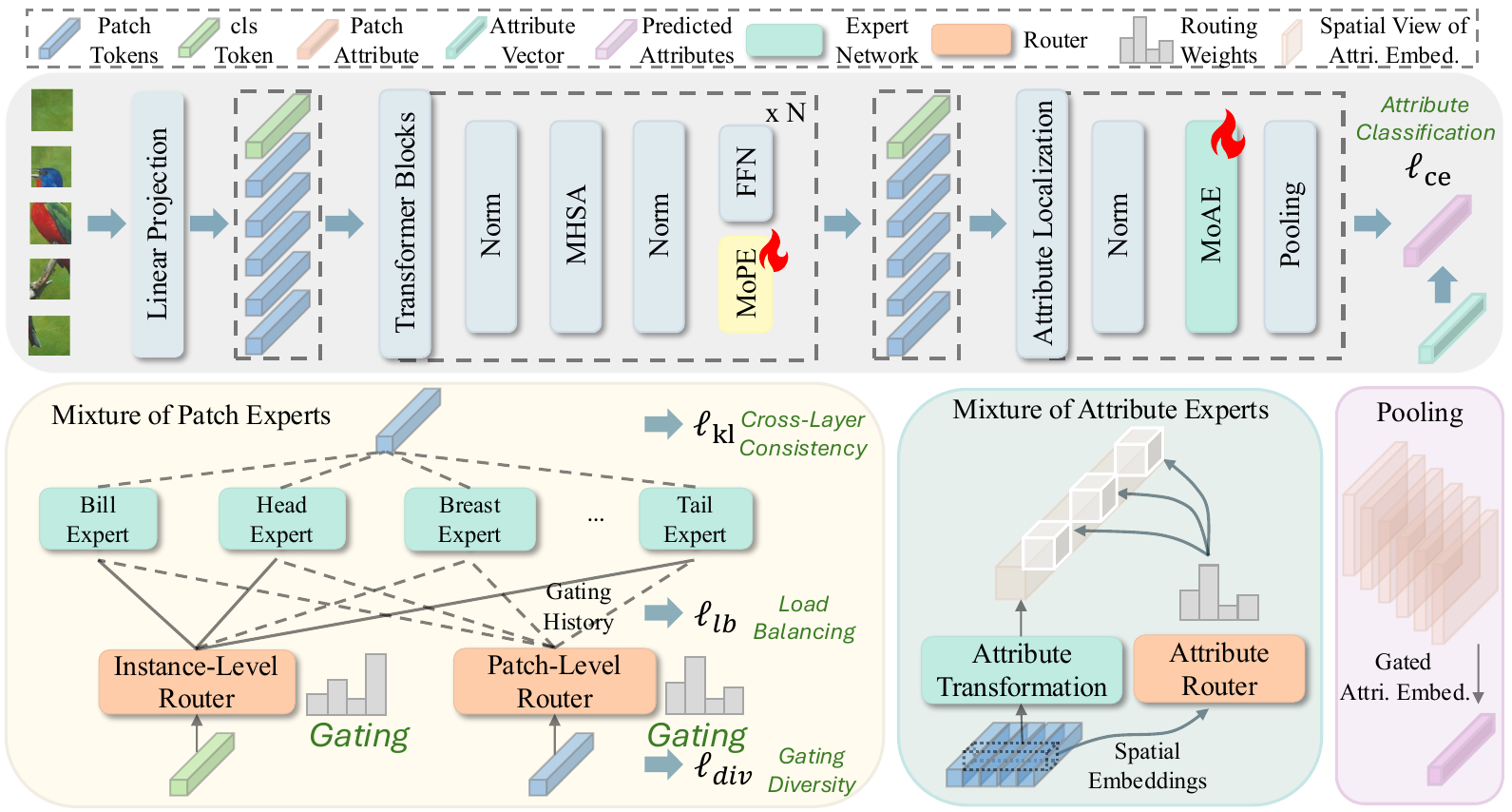}
  \caption{The framework of our proposed ACR for fine-grained zero-shot learning. An image is tokenized and fed through a ViT where a Mixture-of-Patch-Experts (MoPE) adapter is inserted between MHSA and the FFN. A CLS-conditioned instance router activates a few attribute groups, and a patch router then assigns each token to its top experts for sparse, token-adaptive updates. After the final block, an attribute-localization head (Mixture-of-Attribute-Experts) projects tokens to the attribute space and routes dimensions to form a sparse spatial attribute map, which is pooled for classification. Training uses cross-layer gate consistency, load balancing, and token-level diversity to prevent expert collapse and stabilize specialization.}
  \label{fig:arc}
\end{figure*}

\section{Methodology}


\subsection{Problem Formulation}
Let $D^{s}=\bigl\{(\mathbf{x}^{s}_{i},\,y^{s}_{i})\bigr\}_{i=1}^{n_{s}}$
and $ D^{u}=\bigl\{(\mathbf{x}^{u}_{j}, \, y^{u}_{j} ) \bigr \}_{j=1}^{n_{u}}$ denote the \emph{seen} and \emph{unseen} image sets, containing \(n_{s}\) and \(n_{u}\) samples, respectively.  
Their labels come from two disjoint class sets, $C^{s}$ and $C^{u}$. $C^{s}\cap C^{u} = \emptyset $ whose union constitutes the full label space of interest. For every class \(y \in C^{s}\cup C^{u}\) we are given a semantic descriptor, a human‐defined attribute vector, denoted \(\mathbf{a}_{y}\in\mathcal{A}\). During training the model has access \emph{only} to the labelled, attribute‐annotated pairs in \(D^{s}\). No labelled examples from \(C^{u}\) are observed.  The learning objective is therefore to exploit the correspondence between visual samples in \(D^{s}\) and their semantics in \(\mathcal{A}\) so as to construct a classifier that, at test time, can correctly assign any image \(\mathbf{x}\) from the unseen set to one of the classes in \(C^{u}\) without ever having encountered labeled instances of those classes during training.

\subsection{ViT Recap.}  
An input RGB image $\mathbf{I}\!\in\!\mathbb{R}^{H\times W\times C}$ is split into $M$ fixed-size patches $\{\mathbf{p}_{1},\dots,\mathbf{p}_{M}\}$. After flattening, each patch is linearly projected to a $d$-dimensional embedding:
\begin{equation}
\{\mathbf{e}_{1},\dots,\mathbf{e}_{M}\} = \texttt{PatchProj}(\{\mathbf{p}_{1},\dots,\mathbf{p}_{M}\}).
\label{eq:vit_proj}
\end{equation}
A learnable class token $\mathbf{e}_{\langle\mathrm{cls}\rangle}$ is prepended and positional encodings $\mathbf{E}_{\mathrm{pos}}$ are added to form the input sequence for a $L$-layer encoder:
\begin{equation}
\begin{aligned}
\mathbf{X}^{(0)} &= [\mathbf{e}_{\langle\mathrm{cls}\rangle}, \mathbf{e}_{1},\dots,\mathbf{e}_{M}] + \mathbf{E}_{\mathrm{pos}},  \\
\mathbf{X}^{(\ell)} &= \texttt{EncoderLayer}^{(\ell)}(\mathbf{X}^{(\ell-1)}).
\label{eq:vit_tokseq}
\end{aligned}
\end{equation}
The final output is $\mathbf{X}^{(L)} = [\mathbf{h}_{\langle\mathrm{cls}\rangle}, \mathbf{h}_{1},\dots,\mathbf{h}_{M}]$, where $\mathbf{h}_{\langle\mathrm{cls}\rangle}$ is a global summary and $\mathbf{h}_{m}$ retain patch details.

\subsection{Representation Routing with a MoPE}
\label{subsec:mope}

We construct a \textbf{Mixture of Patch Experts} (\textsc{MoPE}) adapter between the Multi-Head Self-Attention (MHSA) block and Feed-Forward Network (FFN) of each ViT block as shown in Figure~\ref{fig:arc}. Instead of updating all patches with the same FFN, we route each token to a small, input-dependent set of expert networks and add their updates residually.
At transformer layer $\ell$, let
\begin{equation}
\mathbf{X}^{(\ell)} = [\mathbf{h}_{\langle\mathrm{cls}\rangle}, \mathbf{h}_{1},\dots,\mathbf{h}_{M}]
\in \mathbb{R}^{(1+M)\times d}.
\end{equation}
\textsc{MoPE} defines $E$ expert networks $\{\mathrm{EXP}_{1},\dots,\mathrm{EXP}_{E}\}$ and activates only the top-$k$ experts ($k\ll E$) per patch.

\vspace{-5pt}
\paragraph{Instance-Level Routing.}
To capture the global information on which attributes the image is biased towards, we propose an instance-level routing mechanism. This is a conventional MoE design choice that purely determines the experts based on image-level information.
The $\mathbf{h}_{\langle\mathrm{cls}\rangle}$ token provides a global image-level summary. We thus use it to determine which experts are globally relevant with an instance router. We keep only the top-$k$ entries of the scores:
\begin{equation}
\mathbf{u}^{\mathrm{logit}} = \mathbf{W}^{I}\mathbf{h}_{\langle\mathrm{cls}\rangle} \in \mathbb{R}^{E},
\label{eq:inst_router}
\end{equation}
where $\mathbf{W}^I$ is the weights of the instance router. 

\vspace{-5pt}
\paragraph{Patch-Level Routing.}
While the instance router provides a global tendency towards experts for the input image, we are particularly interested in learning a patch-level MoE to facilitate the attribute disentangling. With a patch router, each patch token $\mathbf{h}_{m}$ obtains its own routed distribution:
\begin{equation}
\mathbf{v}_{m}^{\mathrm{logit}} = \mathbf{W}^{P}\mathbf{h}_{m}  \in \mathbb{R}^{E}, \quad {\mathrm{for}}~m~{\mathrm{in}}~ \{1, \ldots, M\}.
\label{eq:patch_router}
\end{equation}
We then blend instance and patch evidence via a mixture:
\begin{equation}
{\mathbf{g}}_{m} = (1-\alpha)\,\mathbf{u}^{\mathrm{logit}} + \alpha\,\mathbf{v}_{m}^{\mathrm{logit}},
\label{eq:combined_weights}
\end{equation}
with a mixing coefficient $\alpha\in[0,1]$. 
Keep only the top $k$ experts per token and normalize once with a softmax with a temperature $\tau$, we obtain the routing score:
\begin{equation}
\mathbf{w}_{m} \;=\; \mathrm{softmax}\!\left(\frac{\mathcal{M}_{k}({\mathbf{g}}_{m})}{\tau}\right)
\;\in\; \mathbb{R}^{E}.
\end{equation}
where $\mathcal{M}_{k}$ masks out all but the top-$k$ score entries.

\vspace{-5pt}
\paragraph{Expert Adapters.} 
We construct each expert $k$ with a rank-$r$ Low-Rank Adaptation (LoRA) \cite{hu2022lora} adapter:
\begin{equation}
\begin{aligned}
\operatorname{EXP}_{k}(\mathbf{h})
     &= W_{B}^{(k)}\bigl(W_{A}\mathbf{h}\bigr), \\
W_{A}^{(k)}\!\in\!\mathbb{R}^{d\times r}, \quad
&W_{B}^{(k)}\!\in\!\mathbb{R}^{r\times d},\quad r\!\ll\!d,
\label{eq:expert_def}
\end{aligned}
\end{equation}
where $W_{A}$ casts all tokens into $r$-dimensional vectors and $W_{B}^{(k)}$ is activated based on the routing score $\mathbf{w}_{m}$. 

\vspace{-5pt}
\paragraph{Token Update.}
For patch $m$, let $\mathcal{S}_{m}=\operatorname{TopK}(\mathbf{w}_{m})$ be the indices of its active experts. The MoPE update is the weighted sum of those experts:
\begin{equation}
\Delta\mathbf{h}_{m}
   =\sum_{k\in\mathcal{S}_{m}}
      \frac{\mathbf{w}_{m}[k]}{\sum_{j\in\mathcal{S}_{m}}\mathbf{w}_{m}[j]}
      \;\operatorname{EXP}_{k}(\mathbf{h}_{m}).
\label{eq:token_update}
\end{equation}
The adapted token then follows the standard FFN residual:
\begin{equation}
\mathbf{h}^{\text{out}}_{m}
     =\operatorname{FFN}(\mathbf{h}_{m})\;+\;\Delta\mathbf{h}_{m}.
\label{eq:ffn_out}
\end{equation}
Because only the routed experts are executed, MoPE adds a negligible compute overhead while providing rich, patch-specific capacity.

\subsection{Attribute Localization with MoAE}\label{subsec:attr_loc}
After $L$ transformer layers the MoPE-enhanced outputs are:
\begin{equation}
\mathbf{X}^{(L)}
   = \bigl[\mathbf{h}_{\langle\text{cls}\rangle},
            \mathbf{h}_{1},\dots,\mathbf{h}_{M}\bigr],
\end{equation}
where the CLS token encodes a global summary and each
$\mathbf{h}_{m}\!\in\!\mathbb{R}^{d}$ is a patch-level representation
already refined by its patch specialists (Sec.~\ref{subsec:mope}).  
We next \emph{explicitly localise} class attributes within these patch
tokens so that zero-shot recognition can reason on a sparse,
semantically grounded routing signal.

\vspace{3pt}
\noindent \textbf{Attribute Transformation.}
To project the patch embeddings from the latent dimension to the attribute space, we construct a linear function $\mathrm{AT}\!:\mathbb{R}^{d}\!\to\!\mathbb{R}^{A}$. The attribute embeddings can be obtained by:
\begin{equation}
\begin{aligned}
\mathbf{a}_{m}
   = \mathrm{AT}(\mathbf{h}_{m}), \quad 
\mathbf{A}
   = [\mathbf{a}_{1},\dots,\mathbf{a}_{M}]
   \in\mathbb{R}^{A\times M}.
\label{eq:attr_vector}
\end{aligned}
\end{equation}
The attribute embeddings $\mathbf{A}\in\mathbb{R}^{A\times M} $ can be averaged or maxed over patches to obtain the attribute prediction $\hat{\mathbf{a}}\in\mathbb{R}^A$. In ViT, however, considering the large number of tokens and the small size of patches, the max-pooling operation may be dominated by certain misleading patches, and the mean-pooling may be smoothed out by irrelevant tokens. We thus introduce the following attribute router to localize the relevant patches for each attribute dimension.


\vspace{3pt}
\noindent \textbf{Attribute Router.}
For every attribute embedding $\mathbf{A}_{[a,:]}\in\mathbb{R}^{1 \times M}, a \in \{1,\ldots,A\}$, we predict a sparse one–hot mask
$ \mathbf{f}_{a}\in\{0,1\}^{A}, \; \|\mathbf{f}_{a}\|_{0}=1,$
where $A$ is the number of attributes. The attribute router can produce the routing score:
\begin{equation}
\mathbf{f}_a \;=\; \mathrm{softmax}\!\left(
\frac{\mathcal{M}_{j}(\mathbf{W}^{A}(\mathbf{A}_{[a,:]})}
{\tau}\right)
\;\in\; \mathbb{R}^{M},
\label{eq:attr_router}
\end{equation}
where $\mathcal{M}_{j}$ masks out all but the top-$j$ score entries. We use the straight-through Gumbel trick \cite{huijben2022review} during training and a hard \texttt{argmax} at inference.
We then apply the routing score to the transformed attribute embeddings:
\begin{equation}
\tilde{\mathbf{A}}
   = [\mathbf{A}_{[1,:]}\odot\mathbf{f}_{1},
   \dots,
   \mathbf{A}_{[A,:]}\odot\mathbf{f}_{A}]
   \in\mathbb{R}^{A\times M}.
\end{equation}
A mean-pool over the spatial dimension converts the heat-map into a
global attribute prediction
$\hat{\mathbf{a}}\!=\!\mathrm{mean}_{m}\mathbf{A}[:,m]$.

\subsection{Optimization}
\label{subsec:regularizers}
\paragraph{Classification Loss.}
Given class attributes $\{\mathbf{a}_{c}\}_{c\in {C}^{s}}$, we score seen classes using the attribute prediction $\hat{\mathbf{a}}\in\mathbb{R}^{A}$:
\begin{equation}
s_{c} = \tau\,\langle \hat{\mathbf{a}}, \mathbf{a}_{c} \rangle,
\quad
\mathcal{L}_{\mathrm{cls}} = \mathrm{CE}\big(\mathrm{softmax}(\mathbf{s}),\,y^{s}\big),
\label{eq:class_score}
\end{equation}
where $\tau$ is a temperature and $y^{s}$ is the ground-truth seen label. At test time for ZSL/GZSL we swap in the unseen or the union of seen and unseen attributes.


\vspace{3pt}
\noindent \textbf{Load Balancing Loss.}
Sparse routing scores can cause expert collapse, where a few experts absorb most traffic. To tackle this issue, we measure average usage per expert and penalize dispersion to keep experts active.
For a batch of $B$ images with $M$ non-CLS tokens per image, the average usage at layer $\ell$ is:
\begin{equation}
\bar{\mathbf{w}}^{(\ell)} \;=\; \frac{1}{BM}\sum_{b=1}^{B}\sum_{m=1}^{M} \mathbf{w}^{(\ell)}_{b,m}
\;\in\; \Delta^{E-1}.
\end{equation}
We define normalized usage $U_{e}=E\,\bar{\mathbf{w}}^{(\ell)}_{e}$ so that $\frac{1}{E}\sum_{e}U_{e}=1$ and perfect balance corresponds to $U_{e}\!=\!1$. Let
$\mu_{U}=\frac{1}{E}\sum_{e}U_{e}$ and 
$\sigma_{U}=\sqrt{\frac{1}{E}\sum_{e}(U_{e}-\mu_{U})^{2}}$.
We minimize the load balancing loss via:
\begin{equation}
\mathcal{L}_{\text{lb}}^{(\ell)} 
\;=\; \frac{\mu_{U}}{\sigma_{U}+\varepsilon}
\;=\; \frac{\sqrt{\frac{1}{E}\sum_{e}(U_{e}-\mu_{U})^{2}}}{\frac{1}{E}\sum_{e}U_{e}+\varepsilon},
\label{eq:load_cv}
\end{equation}
with small $\varepsilon\!>\!0$ to stabilize training. 

\vspace{3pt}
\noindent \textbf{Cross-Layer Consistency.}
Routing decisions should reflect a patch prior that is stable across depth. If a token is sent to a “color” specialist early on, later layers should not oscillate toward unrelated experts without evidence. To encourage such stability during representation learning, we regularize layerwise gates toward a per-token consensus.

For a single image with $M$ non\mbox{-}CLS tokens and $L$ MoE layers,
let $\mathbf{w}^{(\ell)}_{m}\in\Delta^{E-1}$ be the routing distribution
over $E$ experts at layer $\ell$ for token $m$.
We define the per-token layerwise average
$ \bar{\mathbf{w}}_{m} \;=\; \frac{1}{L}\sum_{\ell=1}^{L}\mathbf{w}^{(\ell)}_{m}, \qquad \bar{\mathbf{w}}_{m}\in\Delta^{E-1}.$
To encourage stable routing across depth, we penalize the forward KL
from each layer to this average:
\begin{equation}
\mathcal{L}_{\mathrm{cons}}
\;=\;
\frac{1}{M\,L}
\sum_{m=1}^{M}\sum_{\ell=1}^{L}
\mathrm{KL}\!\big(
\mathbf{w}^{(\ell)}_{m}
\;\big\|\;
\bar{\mathbf{w}}_{m}
\big).
\label{eq:consistency_kl_per_image}
\end{equation}

\vspace{3pt}
\noindent \textbf{Diversity Loss.}
Early in training, gates can become overly peaky, which harms exploration and
funnels traffic into a few experts. We promote per-token entropy so that
multiple plausible experts remain active before TopK hardens choices. 
To do so, we maximize the routing score entropy:
\begin{equation}
\begin{aligned}
\mathcal{L}_{\text{div}} \;=\; -\;\frac{1}{M\,L}\sum_{\ell=1}^{L_{}}\sum_{m=1}^{M}
H\!\big(\mathbf{w}^{(\ell)}_{m}\big) , \\
H(\mathbf{w}) = -\sum_{e=1}^{E} w_{e}\,\log\!\big(w_{e}+\varepsilon\big),
\label{eq:diversity_no_batch}
\end{aligned}
\end{equation}
with a small $\varepsilon$ for numerical stability.
Note that $\mathcal{L}_{\text{lb}}$ (expert usage balance) acts at the {marginal} expert level, whereas $\mathcal{L}_{\text{div}}$ preserves {token-level} exploration.

\vspace{3pt}
\noindent \textbf{Full Objective}
\label{subsec:full_objective}
The training loss combines attribute supervision, seen-class recognition, and routing regularizer:
\begin{equation}
\mathcal{L}_{\mathrm{total}}
= \underbrace{\mathcal{L}_{\mathrm{cls}}}_{\text{Class CE}}
+\lambda_{1} \mkern-20mu\mathord{\underbrace{\mathcal{L}_{\mathrm{lb}}}_{\text{LoadBalancing}}} \mkern-15mu
\mathord{+} \lambda_{2}\!\underbrace{\mathcal{L}_{\mathrm{cons}}}_{\text{Consistency}}
\mkern-10mu
\mathord{+} \lambda_{3}\underbrace{\mathcal{L}_{\mathrm{diver}}}_{\text{Diversity}},
\label{eq:total_loss}
\end{equation}
where $\lambda_1, \lambda_2$ and $\lambda_3$ are the loss coefficients.

\section{Experiments}
\subsection{Setup and Implementation}
We evaluate our approach on three widely used zero-shot benchmarks, including CUB \cite{wah2011caltech}, AWA2 \cite{lampert2013attribute}, and SUN\cite{patterson2014sun}. CUB contains 11,788 bird images from 200 categories, each described by 312 fine-grained attributes. AWA2 offers 37,322 images spanning 50 animal species with 85 attributes, while SUN (Patterson et al., 2014) comprises 14,340 images from 717 scene classes annotated by 102 attributes. We adopt the standard splits of  \cite{xian2018feature}, yielding 150/50, 40/10 and 645/72 seen-to-unseen class partitions for CUB, AWA2 and SUN, respectively. Performance is reported as top-1 accuracy under both the conventional zero-shot setting, where evaluation is restricted to unseen classes, and the generalised setting, which measures accuracy on seen (S) and unseen (U) test samples separately. To balance these two numbers we follow \cite{xian2018zero} and quote their harmonic mean $H = (2\times S\times U)/(S + U)$, a single score that rewards models which perform well on both parts of the label space.
Our architecture builds on the ViT-Base backbone \cite{touvron2021training}. We initialise the network with ImageNet-1k pretrained weights, following the standard practice adopted by recent ViT-based zero-shot methods.

\begin {table*}[t]
\caption {Comparison of state-of-the-art ZSL and GZSL models on three benchmark datasets. For ZSL, results are reported as average top-1 accuracy (T1). For GZSL, we report top-1 accuracy for unseen (U) and seen (S) classes, along with their harmonic mean (H). The best and second-best results are highlighted in \textbf{bold} and \underline{underlined}, respectively.
}
\centering
\begin{tabular}[t]{@{~}cl@{~}c@{~}|cccc|cccc|cccc}
\hline
  \multirow{2}{*}{}     & & & \multicolumn{4}{c|}{\textbf{CUB}}  &  \multicolumn{4}{c|}{\textbf{AwA2}}   & \multicolumn{4}{c}{\textbf{SUN}}\\ 
  
\hhline{|~~~|------------|}
& \multirow{-2}{*}{\textbf{Methods}} & \multirow{-2}{*}{\textbf{Venue}} & \multicolumn{1}{c|}{\textit{T1}} &\textit{U} & \textit{S} & \textit{H}  & \multicolumn{1}{c|}{\textit{T1}} & \textit{U} & \textit{S} & \textit{H}  & \multicolumn{1}{c|}{\textit{T1}} & \textit{U} & \textit{S} &        \textit{H} \\
\hline
\multirow{10}{*}{\rotatebox{270}{ResNet101}}    
& TF  \cite{narayan2020latent}&ECCV'20   
&\multicolumn{1}{c|}{64.9}      & 52.8      & 64.7      & 58.1
&\multicolumn{1}{c|}{{72.2}}    & 59.8      & 75.1      & 66.6  
& \multicolumn{1}{c|}{66.0}     & 45.6      & 40.7      & 43.0\\

&RFF  \cite{han2020learning}&   CVPR'20     
&\multicolumn{1}{c|}{-}    & 52.6          & 56.6    & 54.6
&\multicolumn{1}{c|}{-}    & 59.8          & 75.1    & 66.5   
&\multicolumn{1}{c|}{-}    & 45.7          & 38.6    & 41.9\\ 
&
APN       \cite{xu2020attribute}&   NeurIPS'20    
&\multicolumn{1}{c|}{72.0} & 65.3          & 69.3    & 67.2
&\multicolumn{1}{c|}{68.4} & 56.5          & 78.0    & 65.5
&\multicolumn{1}{c|}{61.6} & 41.9          & 34.0    & 37.6\\


&SDGZSL \cite{chen2021semantics}    &  ICCV'21   
&\multicolumn{1}{c|}{75.5}    & 59.9        & 66.4      & 63.0
&\multicolumn{1}{c|}{72.1}    & 64.6        & 73.6      & 68.8    
&\multicolumn{1}{c|}{62.4}    & 48.2        & 36.1      & 41.3\\

&Dis-VAE \cite{li2021generalized}    &  AAAI'21   
&\multicolumn{1}{c|}{-}    & 51.1        & 58.2      & 54.4
&\multicolumn{1}{c|}{-}    & 56.9        & 80.2      & 66.6    
&\multicolumn{1}{c|}{-}    & 36.6        & 47.6      & 41.4\\

& GEM  \cite{liu2021goal}&   CVPR'21   
&\multicolumn{1}{c|}{77.8} & 64.8          & 77.1     & 70.4
&\multicolumn{1}{c|}{67.3} & 64.8          & 77.5     & 70.6
&\multicolumn{1}{c|}{62.8} & 38.1          & 35.7     & 36.9\\

&
HSVA \cite{chen2021hsva}      &NeurIPS'21
&\multicolumn{1}{c|}{62.8}    & 52.7        & 58.3      & 55.3
&\multicolumn{1}{c|}{-}       & 59.3        & 76.6      & 66.8
&\multicolumn{1}{c|}{63.8}    & 48.6        & 39.0      & 43.3 \\
&
MSDN  \cite{chen2022msdn}&      CVPR'22   
&\multicolumn{1}{c|}{76.1} & 68.7          & 67.5     & 68.1
&\multicolumn{1}{c|}{70.1} & 62.0          & 74.5     & 67.7    
&\multicolumn{1}{c|}{65.8} & \underline{52.2} & 34.2 & 41.3 
\\
&ICCE \cite{kong2022compactness}    &  CVPR'22 
&\multicolumn{1}{c|}{71.6}    & 67.3        & 65.5      & 66.4
&\multicolumn{1}{c|}{{78.4}} & 65.3  & 82.3  & 72.8
&\multicolumn{1}{c|}{-}       & -           & -         & -   \\

& 
ICIS \cite{christensen2023image}  & ICCV’23 
& \multicolumn{1}{c|}{60.6} & 45.8 & 73.7 & 56.5 
& \multicolumn{1}{c|}{64.6} & 35.6 & \textbf{93.3} & 51.6
& \multicolumn{1}{c|}{51.8} & 45.2 & 25.6 & 32.7 \\ 

\hline
    
\multirow{7}{*}{\rotatebox{270}{ViT-based}}

&CLIP  \cite{radford2021learning}&      ICML'21      
&\multicolumn{1}{c|}{-} & 55.2          & 54.8     & 55.0
&\multicolumn{1}{c|}{-} & -             & -        & -   
&\multicolumn{1}{c|}{-} & -             & -        & - 
\\

&CoOp  \cite{zhou2022learning}&      IJCV'22     
&\multicolumn{1}{c|}{-} & 49.2          & 63.8     & 55.6
&\multicolumn{1}{c|}{-} & -             & -        & -   
&\multicolumn{1}{c|}{-} & -             & -        & - 
\\

&I2DFormer  \cite{naeem2022i2dformer}&      NeurIPS'22
&\multicolumn{1}{c|}{45.4} & 35.3          & 57.6     & 43.8
&\multicolumn{1}{c|}{76.4} & 66.8          & 76.8     & 71.5   
&\multicolumn{1}{c|}{-} & -             & -        & - 
\\

&
I2MV  \cite{naeem2023i2mvformer}&      CVPR'23    
&\multicolumn{1}{c|}{42.1} & 32.4          & 63.1     & 42.8
&\multicolumn{1}{c|}{\underline{73.6}} & {66.6}          & 82.9     & 73.8    
&\multicolumn{1}{c|}{-}    & -             & -        & - 
\\
&
DUET  \cite{chen2023duet} &      AAAI'23      
&\multicolumn{1}{c|}{72.3} & 62.9 & 72.8     & 67.5
&\multicolumn{1}{c|}{69.9} & 63.7          & 84.7     & 72.7    
&\multicolumn{1}{c|}{64.4} & 45.7          & 45.8        & 45.8 
\\

&ZSLViT  \cite{chen2024progressive} & CVPR'24  
&\multicolumn{1}{c|}{\underline{78.9}}    
& \underline{69.4}          
& \underline{78.2}              
& \underline{73.6}
&\multicolumn{1}{c|}{70.7} 
& \underline{66.1}          
& 84.6      
& \underline{74.2}
&\multicolumn{1}{c|}{\underline{68.3}}   
& 45.9          
& \underline{48.4}          
& \underline{47.3} 
\\
&VSPCN \cite{jiang2025visual} & CVPR'25 & \multicolumn{1}{c|}{80.6} & 72.8 & 78.9 & 75.7 & \multicolumn{1}{c|}{76.6} & 71.8 & 84.3 & 77.6 & \multicolumn{1}{c|}{75.3} & 59.4 & 49.1 & 53.8 
\\
&
SVIP \cite{chen2025svip} & ICCV'25
& \multicolumn{1}{c|}{{79.8}} 
& {72.1} 
& {78.1} 
& {75.0}   
& \multicolumn{1}{c|}{{69.8}} 
& 65.4 
& \underline{87.7} 
& {74.9}
&\multicolumn{1}{c|}{{71.6}} 
& {53.7} 
& {48.0} 
& {50.7}
\\
\hhline{|~|--------------|}
&
\textbf{ACR}(ours) & 
& \multicolumn{1}{c|}{\textbf{80.9}} & \textbf{72.8} & \textbf{82.2} & \textbf{77.2}   
& \multicolumn{1}{c|}{\textbf{79.1}} & \textbf{74.1} & {86.3} & \textbf{79.7}
&\multicolumn{1}{c|}{\textbf{76.5}} & \textbf{60.0} & \textbf{51.0} & \textbf{55.1}
\\
\hline
\end{tabular}
\label{gzslperoformance}
\end {table*}

\subsection{Quantitative Results}
The ResNet101-based methods established a strong but clearly saturated baseline. On CUB, the best CNN harmonic mean is GEM \cite{liu2021goal} with 70.4\%, whereas ACR reaches 77.2\%, which achieves an absolute gain of 6.8\%. On AwA2, ICCE \cite{kong2022compactness} achieves 72.8\% on H, whereas ACR pushes this to 79.7\% (+6.9\%). SUN is the most challenging due to diffuse, scene-level attributes: the strongest CNN H hovers around 43.0\% (TF \cite{narayan2020latent}) with similar values for MSDN \cite{chen2022msdn}, yet ACR delivers 55.1\%, improving the seen–unseen balance by more than twelve points. These margins indicate that modeling attribute structure during representation learning, not just at the classifier, addresses the failure modes of global CNN features, where subtle attributes are easily swamped by context.

Among ViT backbones, ACR consistently advances the state of the art. On CUB, it attains the top ZSL accuracy (T1=80.9\%) and the best GZSL harmonic mean (H=77.2\%), matching the strongest prior unseen score (U=72.8\% from VSPCN \cite{jiang2025visual}) while substantially lifting seen accuracy to S=82.2\% (vs. 78.9\% for VSPCN \cite{jiang2025visual}). On AwA2, ACR also sets the new ZSL peak (T1=79.1\%) and raises H to 79.7\%, surpassing the previous best 77.6\% (VSPCN), with U=74.1\% and S=86.3\% indicating balanced calibration instead of the “seen-only” bias observed in ICIS \cite{christensen2023image} (S=93.3\%, U=35.6\%, H=51.6\%). On SUN, ACR improves ZSL to 76.5\% and, crucially, delivers U=60.0\% and lifts H to 55.1\% over the prior best around 50–54\%. The SUN gains particularly indicate that when attributes are sparse and noisy, routed specialists preserve weak but decisive cues that generic ViT heads tend to wash out.

\vspace{3pt}
\noindent \textbf{Ablation Study.}
We conduct an ablation study on CUB and AwA2 to validate the effectiveness of each component. As shown in Table \ref{Tab:ablations}, ACR improves over a plain ViT with an attribute head across both datasets. Removing the cross-layer consistency term (w/o $\mathcal{L}_{cons}$) weakens routing stability and reduces H on CUB/AwA2 (-2.6/-6.4), showing that aligning gates across depth helps specialists remain coherent. Disabling load balancing (w/o $\mathcal{L}_{lb}$) causes expert collapse and hurts H notably on CUB (0.9) and AwA2 (-1.9), confirming the need to spread traffic. Turning off the token-level diversity (w/o $\mathcal{L}_{div}$) slightly degrades H (-1.7/-3.1), indicating early exploration prevents overly peaky gates. The largest drop comes from removing the patch router (w/o Patch Router), which forces uniform expert updates and erases most gains (H -3.1/-3.3), demonstrating that token-adaptive routing is essential. Overall, each component contributes, with routing (patch-level) and cross-layer consistency being the primary drivers of the final performance.

\begin{table}[t]
\centering
\caption{Component Analysis of ACR on CUB and AwA2. }
\label{Tab:ablations}
\scalebox{0.87}{
\begin{tabular}{ l @{~} | @{~~} c @{~~} c @{~~} c @{~~} c @{~~} | @{~~} c @{~~} c @{~~} c @{~~} c @{~~}  }
    \hline
    \multirow{2}{*}{Method}  
    & \multicolumn{4}{c}{CUB} 
    & \multicolumn{4}{c}{AwA2} \\ 
    
    \hhline{~--------}
     & $T1$ & $U$ & $S$ & $H$ & $T1$ & $U$ & $S$ & $H$  \\ \hline \hline
    
    Baseline(ViT)   
    & 76.8  & 59.8  & 68.4  & 63.8    
    & 61.4  & 58.0  & 81.6  & 67.8   \\
                    
    ACR (full)                   & \textbf{80.9}  & \underline{72.8}  & \textbf{82.2}  &        \textbf{77.2}  
                                    & \textbf{79.1}  & \textbf{74.1}  & {86.3}  & \textbf{79.7}   
                                    \\
                                    
    -- w/o $\mathcal{L}_{cons}$ 
                                    & 78.8  & 71.7  & 77.8  & 74.6  
                                    & 71.5  & 62.0  & \textbf{89.6}  & 73.3  
                                     \\

    -- w/o $\mathcal{L}_{lb}$ 
                                    & 77.7  & \textbf{74.8} & 77.8 & \underline{76.3} 
                                    & \underline{75.7}  & \underline{73.2}  & 83.9 & 78.2     
                                     \\
                                  
    -- w/o $\mathcal{L}_{div}$ 
                                    & \underline{80.0} & 71.2 & \underline{80.3} & 75.5 
                                    & 75.1 & 71.7 & 82.5 & 76.7 
                                    \\
    -- w/o P. Routing 
                                    & 76.9 & 70.2 & 78.4 & 74.1
                                    & 72.4 & 66.0 & \underline{89.2} & 75.9
                                    \\
        \bottomrule
\end{tabular}}
\end{table}

\begin{figure*}[t]
	\centering
	\includegraphics[width=1\textwidth]{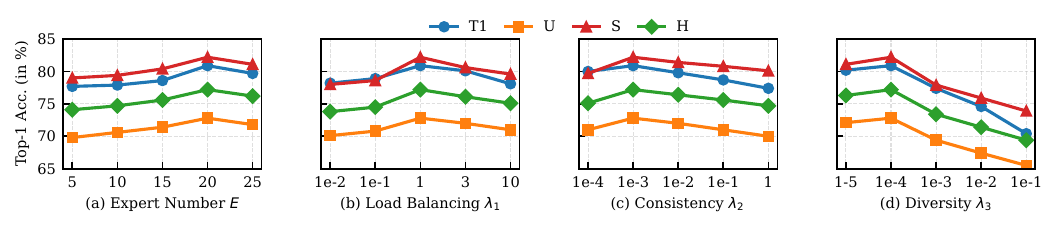}
    \vspace{-15pt}
	\caption{Hyper-parameter sensitivity on CUB dataset.}
	\label{FIG:hyper}
    \vspace{-5pt}
\end{figure*}

\vspace{3pt}
\noindent \textbf{Hyper-Parameter Sensitivity.}
Figure~\ref{FIG:hyper} analyses four key hyperparameters on CUB. 
(a) Varying the number of experts $E$ shows a clear accuracy trend, where performance improves as $E$ increases and peaks around $E{=}20$, after which marginal benefits vanish and H drops slightly at $E{=}25$, likely due to increased routing variance. 
(b) The load-balancing weight $\lambda_{1}$ exhibits a broad optimum in the $1\!\sim3$ range and too little regularization (1e-2) allows expert collapse (lower U/H), while too much ($10$) over-constrains routing. 
(c) The cross-layer consistency weight $\lambda_{2}$ is most helpful at 1e-3 $\sim$ 1e-2, where it stabilizes specialist identities across depth and raises both U and H. Excessively weak or strong values under- or over-constrain the gates. 
(d) The diversity weight $\lambda_{3}$ benefits early exploration when set near 1e-4, but when pushing it to 1e-3 makes gates overly diffuse and reduces S/H. 

\begin{figure}[t]
	\centering
	\includegraphics[width=1\columnwidth]{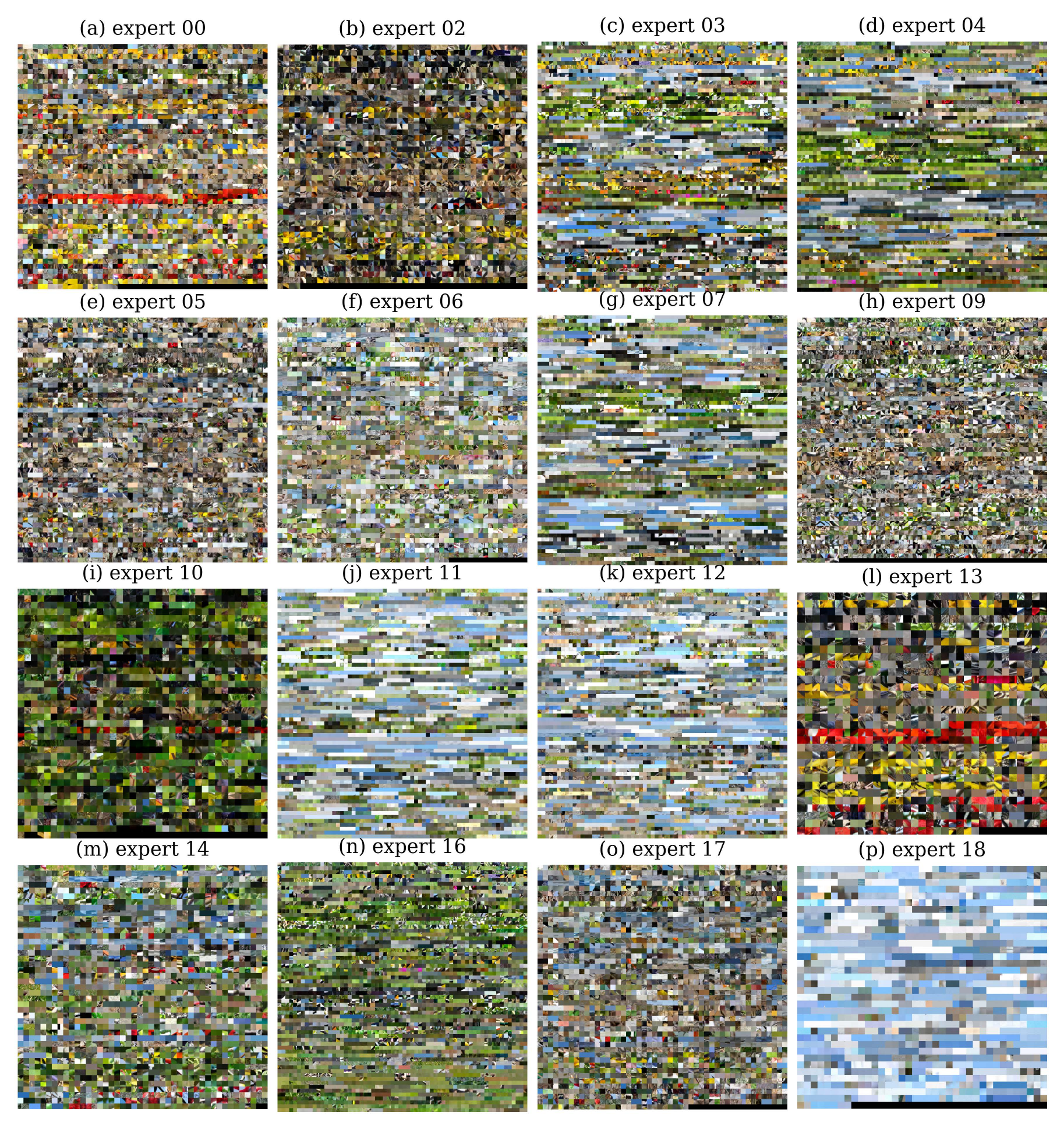}
        \vspace{-15pt}
	\caption{Expert-specific patch collages illustrating the visual patterns captured by the MoPE. Each block aggregates the top-ranked patches for a single expert, arranged row-wise by confidence.} 
	\label{FIG:expert_patches}
    \vspace{-5pt}
\end{figure}

\begin{figure}[t]
	\centering
	\includegraphics[width=1\columnwidth]{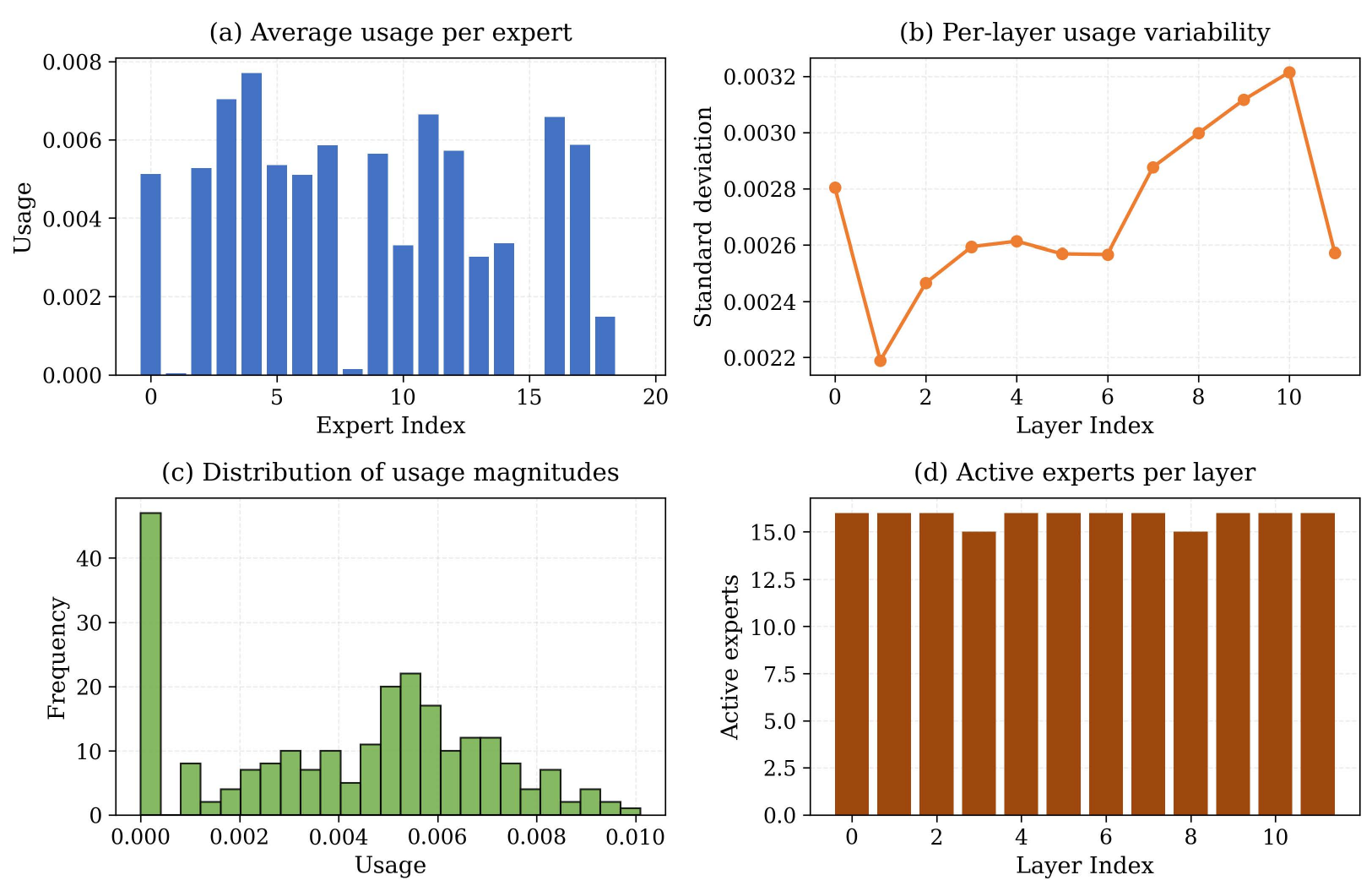}
        \vspace{-10pt}
    	\caption{Expert utilization analysis summarizing routing behaviour across the MoAE. (a) mean routing probability allocated to each expert, (b) variability of expert usage per transformer layer, (c) histogram of usage magnitudes, (d) count of experts activated.}
	\label{FIG:expert_uti}
\end{figure}

\subsection{Qualitative Study}

\begin{figure*}[t]
	\centering
	\includegraphics[width=0.98\textwidth]{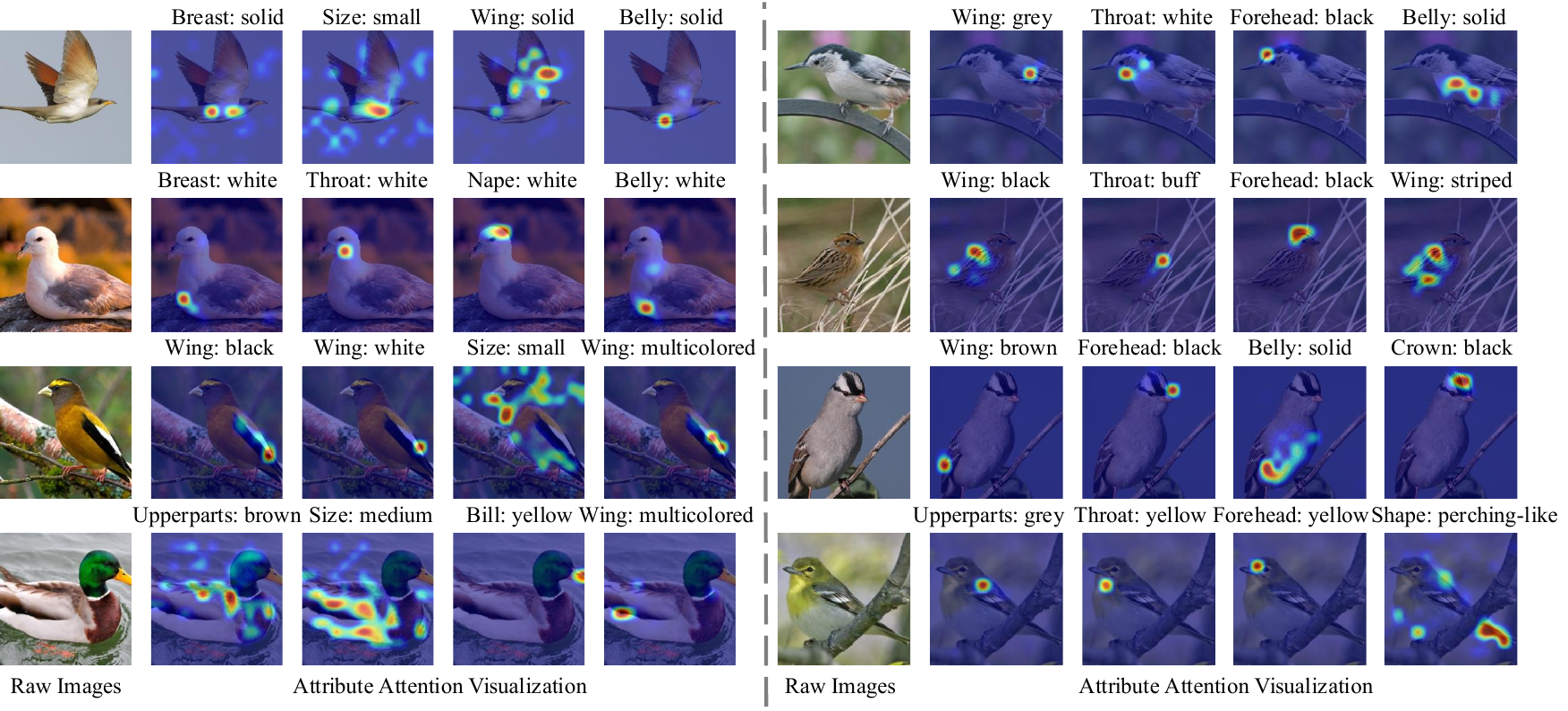}
        \vspace{-5pt}
	\caption{Visualization of attribute localization. For each image of the left tile in each group, we visualize the patch-wise attribute activations produced by the attribute localization head.  Titles above maps denote the routed attribute, e.g., \textit{breast: solid}, \textit{throat\_color: white},.  Warm colors indicate higher activation.  ACR consistently grounds colors on the corresponding parts (head, throat, wings), highlights size on the full silhouette, and localizes pattern on textured regions (belly, wings), even under pose variation and clutter.}
	\label{FIG:attribute_heatmap}
    \vspace{-5pt}
\end{figure*}


\begin{figure}[t]
    \centering
    \begin{subfigure}[b]{0.23\textwidth}
        \centering
        \includegraphics[width=\textwidth]{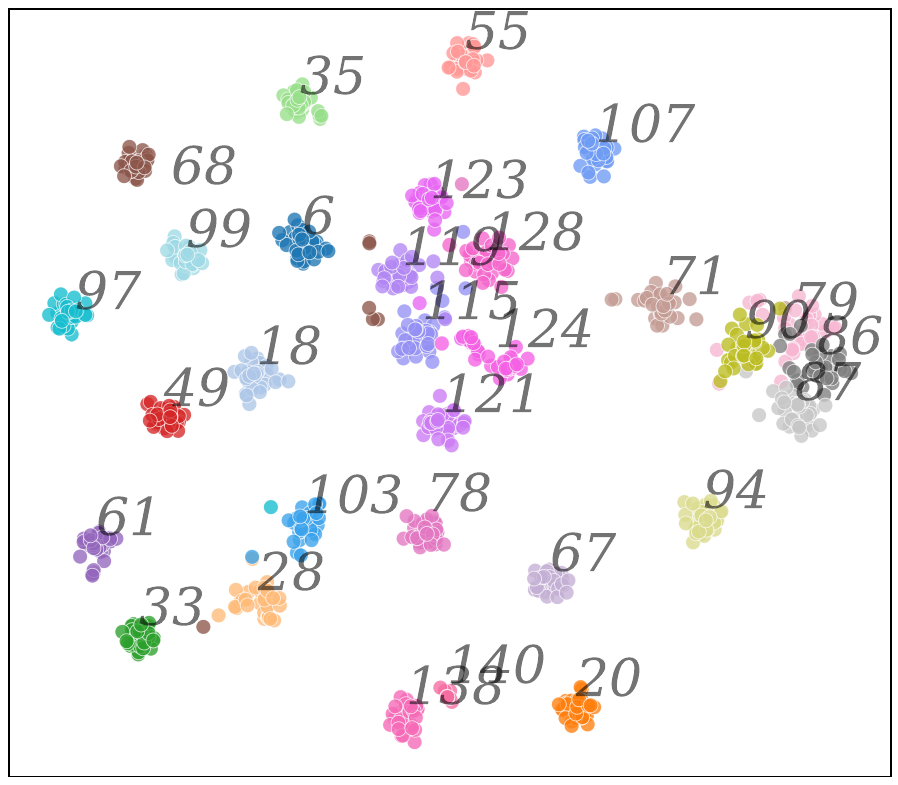}
        \caption{ACR}
        \label{fig:cub_tsne}
    \end{subfigure}
    \hfill
    \begin{subfigure}[b]{0.23\textwidth}
        \centering
        \includegraphics[width=\textwidth]{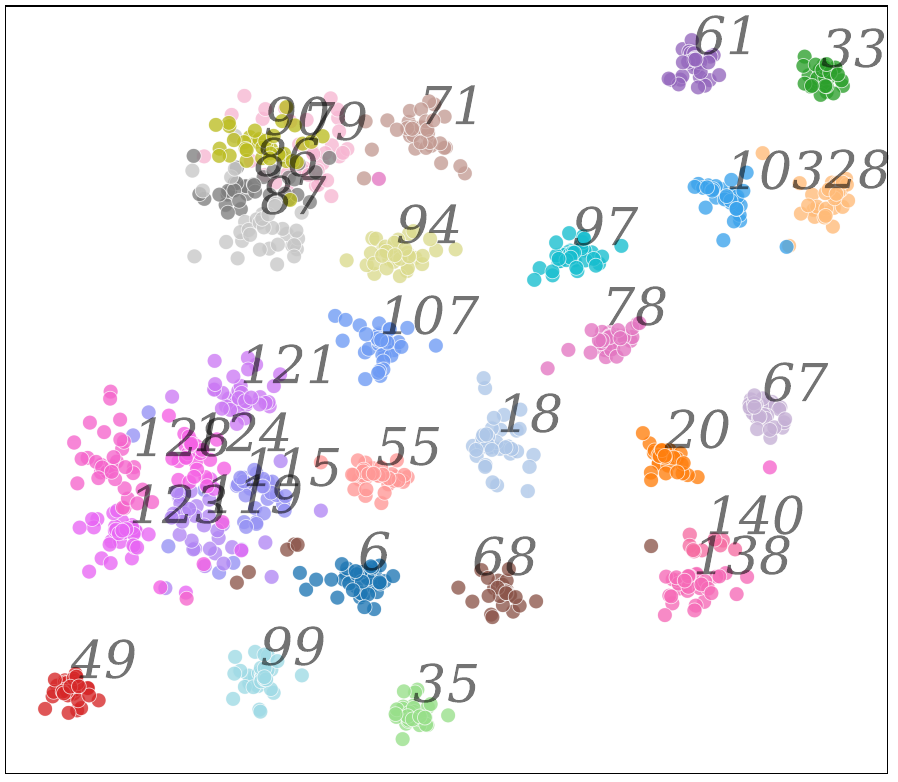}
        \caption{Vanilla ViT}
        \label{fig:awa2_tsne}
    \end{subfigure}
    \caption{t-SNE visualizations of the learned representations on CUB unseen classes. Colors represent different classes and numbers are the class indices.}
    \label{fig:tsne_compare}
    \vspace{-5pt}
\end{figure}

\noindent \textbf{Expert Utilization Analysis.}
We examine the routing behaviour of the MoAE using the summary in Figure \ref{FIG:expert_uti}. (a) shows that token-to-expert assignments remain relatively balanced. Although a few experts dominate, the long tail indicates that most experts contribute meaningfully. It avoids the single expert collapse problem that is often observed in naïvely trained MoEs. (b) reveals the depth-wise dynamics. Early layers exhibit lower variance while deeper layers display more pronounced fluctuations. This suggests that specialization intensifies as features become more semantic. (c) confirms the overall sparsity of the router. The usage histogram is sharply peaked near zero, which means the model activates experts only when needed. This helps control computation. Finally, (d) reports the number of active experts per layer under a 0.1\% threshold. Every layer keeps a broad set of experts engaged, typically 12–15 experts. This reinforces that the learned routing preserves diversity even when only a subset is required for each token. Together, these observations indicate that the proposed training strategy for ACR achieves the desired balance between specialization and load balancing, so that experts are used sparsely yet consistently. Overall, the visualization of the expert utilization provides both interpretability and improved generalization.

\vspace{3pt}
\noindent \textbf{What do Experts Learn?}
Figure~\ref{FIG:expert_patches} visualizes the inductive biases captured by each expert via patch collages. 
For every expert $e$, we gather the top-$K$ routed patches across the test set using the post–softmax gating scores, crop the original $14\times14$ tokens, and tile them row–wise by confidence. 
Distinct, stable patterns emerge where some experts focus on {plumage textures} like fine mottling and feather barbs, others prefer solid color fields like breast/wing blocks, {linear structures} like feather edges, beak rims or background textures like foliage, sky–water bands. 
As the regularization of the consistency loss, these specializations are consistent across layers and images, which indicates that routing separates attribute groups rather than class identities.

\vspace{3pt}
\noindent \textbf{Qualitative Localization.}
Figure~\ref{FIG:attribute_heatmap} shows spatial attribute maps produced by MoAE. Given the routed token embeddings, the attribute head projects each patch to the attribute space and keeps only the selected dimension. The resulting scores are normalized and overlaid as heatmaps. 
Across diverse species and poses, color attributes peak on the correct parts, e.g., {forehead: black}, {throat: white}), pattern attributes focus on belly/wing textures, and {size} highlights the global silhouette rather than background. 
We found that the localization can be very accurate in similar positions, such as the crown and forehead. 
These results indicate that the routed experts preserve attribute-wise structure throughout the backbone and yield interpretable, part-aware evidence for recognition.

\vspace{5pt}
\noindent \textbf{t-SNE Visualization of Unseen-Class Structure.}
To qualitatively assess representation quality, we project global image embeddings from CUB unseen classes into 2D using t-SNE. For ACR we use the pooled attribute vector after the attribute-localization head. For the vanilla ViT baseline, we use the final CLS embedding. Figure \ref{fig:tsne_compare} shows that ACR on the left forms tighter and well-separated clusters with clearer inter-class margins and markedly lower intra-class variance. In contrast, vanilla ViT on the right exhibits overlapping clusters and scattered samples, indicating residual attribute entanglement. These geometric differences align with our quantitative gains on unseen accuracy and harmonic mean.



\section{Conclusion and Limitations}
We presented ACR, a ViT-based mixture-of-experts that routes tokens to attribute specialists during representation learning, yielding sparse, interpretable attribute maps. A dual-level router (instance and patch) activates lightweight LoRA experts only when needed, preserving compute while improving attribute alignment. On CUB, AWA2, and SUN, MoAE achieves state-of-the-art ZSL/GZSL performance, demonstrating that early, specialist-driven disentanglement outperforms post-hoc refinement. 

In ACR, limitations and opportunities remain. Although our regularizers mitigate expert collapse, extreme class imbalance or long-tail attribute distributions can still bias routing. Our current specialization is learned implicitly, thus incorporating stronger semantic priors, e.g., attribute groupings, could accelerate convergence and sharpen specialization. Finally, while we focused on attribute-vector supervision, the same routed backbone could be paired with language models and richer textual prompts to enable unified multimodal understanding and generation.



{
    \small
    \bibliographystyle{ieeenat_fullname}
    \bibliography{main}
}


\end{document}

%% file: math.tex
\usepackage{amsmath,amsfonts,bm}

\def\eqref#1{equation~\ref{#1}}

\def\1{\bm{1}}

\DeclareMathAlphabet{\mathsfit}{\encodingdefault}{\sfdefault}{m}{sl}
\SetMathAlphabet{\mathsfit}{bold}{\encodingdefault}{\sfdefault}{bx}{n}